\documentclass[runningheads]{llncs}
\usepackage[T1]{fontenc}
\usepackage{graphicx}
\usepackage{microtype}
\usepackage{algpseudocode}
\usepackage[ruled, noline,noend]{algorithm2e}
\usepackage{amsmath}
\usepackage{amssymb}
\usepackage{mathtools}
\usepackage{booktabs,siunitx}
\usepackage{tablefootnote}
\usepackage{subcaption}
\usepackage{url}

\begin{document}
\title{Extracting Moore Machines from Transformers using Queries and Counterexamples}
\titlerunning{Extracting Moore Machines from Transformers}
%
\author{Rik Adriaensen\orcidID{0009-0008-7863-4034} \and
Jaron Maene\orcidID{0000-0001-9474-6150}}
\authorrunning{R. Adriaensen and J. Maene}
%
\institute{KU Leuven, Departement of Computer Science, Leuven, Belgium
\email{\{rik.adriaensen,jaron.maene\}@kuleuven.be}
}
\maketitle              
\begin{abstract}
Fuelled by the popularity of the transformer architecture in deep learning, several works have investigated what formal languages a transformer can learn from data. Nonetheless, existing results remain hard to compare due to methodological differences. To address this, we construct finite state automata as high-level abstractions of transformers trained on regular languages using queries and counterexamples. Concretely, we extract Moore machines, as many training tasks used in literature can be mapped onto them.
We demonstrate the usefulness of this approach by studying positive-only learning and the sequence accuracy measure in detail.
\end{abstract}

\section{Introduction}

Transformers \cite{vaswani2017attention} have become a dominant architecture in natural language processing and deep learning more generally. This explains the increase of interest in characterising the expressive power of this architecture. Recently, significant effort has been devoted to proving what formal languages can or cannot be \emph{expressed} by the transformer architecture \cite{ackerman2020survey,merrill2021formal,Strobl2023TransformersAR}. 

The \emph{trainability}---what formal languages transformers can learn from data---remains less well understood. Indeed, the fact that a transformer can express a formal language does not necessarily mean it can also learn said language from data.
Consider the Parity language, for example, which consists of all binary sequences with an odd number of ones. Although Chiang et al. \cite{chiang2022overcoming} proved that soft attention transformers can express this language, experimental studies found that transformers fail to learn the Parity language from data \cite{anil2022exploring,deletang2023neural,liu2023transformers}.

Empirically studying the trainability of transformers is challenging, however. Many possible choices for the specific training setup make it hard to draw consistent conclusions \cite{bhattamishra2020ability,liu2023transformers,yao2021self-membership}. Moreover, small changes in the transformer architecture can significantly impact what languages can or cannot be expressed \cite{Strobl2023TransformersAR}. For example, transformers can no longer express Parity when masked hard attention is used instead of soft attention \cite{yang2023maskedht}.

To study the trainability of regular languages, we propose a method to \emph{extract finite state automata from transformers}. Inspired by prior work that extracts deterministic finite automata from recurrent neural networks \cite{weiss2018extracting}, our method uses an extension of the classical $L^*$ algorithm \cite{angluin1987lstar} to recover the automaton simulated by a transformer. Experimentally, we show that the extracted automata faithfully model the behaviour of transformers that recognise a regular language.

We demonstrate the use of our extraction method by studying the influence of the choice of training task on trainability.
Our extraction method accommodates these different training tasks, as they can be seen as simulating a specific Moore machine.
In our experiments, we find that transformers do not model garbage states when trained on purely positive examples and that the prevalent sequence accuracy measure can report a perfect score when the transformer has learnt the wrong language. 

In summary, the contribution of this paper is: (1) a method for extracting Moore machines from transformers and (2) applying this method to identify two common flaws in trainability studies.

\section{Preliminaries}

We assume basic familiarity with transformers \cite{vaswani2017attention}. In this section, we briefly introduce regular languages and finite state automata. We refer to Hopcroft et al. \cite{hopcroft2006automata} for a more thorough introduction.

We write $\Sigma$ or $\Gamma$ for a finite set of symbols, also called an \emph{alphabet}. The \emph{Kleene closure} $\Sigma^*$ of an alphabet $\Sigma$ is the set of all finite sequences using the symbols of $\Sigma$. For example, the binary alphabet $\Sigma=\{0, 1\}$ has as closure $\Sigma^* = \{ \epsilon, 0, 1, 00, 01, 10, 11, \dots \}$. We denote the empty sequence with $\epsilon$ and the empty set with $\emptyset$. A \emph{language} $L$ over the alphabet $\Sigma$ is a subset of $\Sigma^*$.

\begin{definition} \cite{hopcroft2006automata}
  The \emph{regular languages} are those subsets of $\Sigma$ obtained by closing the empty set and the singleton sets containing a symbol from $\Sigma \cup \{\epsilon\}$ over union, concatenation, and the Kleene operator.
\end{definition}

\noindent
The star-free languages are a subset of the regular languages constructable without using Kleene closure.

\begin{definition} \cite{barrington1992regular}
  The \emph{star-free regular languages} are those subsets of $\Sigma$ obtained by closing the empty set and the singleton sets containing a symbol from $\Sigma \cup \{\epsilon\}$ over union, concatenation, and complement.
\end{definition}

\noindent
For example, the language containing no consecutive zeros can be constructed as $\overline{\Sigma^*00\Sigma^*}$ and, as $\Sigma^* = \overline{\emptyset}$, is therefore star-free.

\begin{definition} A \emph{deterministic finite automaton} (DFA) is a tuple $(Q, \Sigma, \delta, q_0, F)$, such that
$Q$ is a finite set of states,
 $\Sigma$ is the input alphabet,
 $\delta:  Q \times \Sigma \to Q$ is the transition function,
 $q_0 \in Q$ is the start state, and
$F \subset Q$ is the set of final states.

\end{definition}

Given a sequence $s \in \Sigma^*$, a symbol $\sigma \in \Sigma$, and a state $q \in Q$, we can define the extended transition function $\hat\delta: Q \times \Sigma^* \to Q$ as $\hat\delta(q, \sigma . s) = \hat\delta(\delta(q, \sigma), s)$, with $\hat\delta(q,\epsilon) = q$ in the base case. We say that a DFA \emph{accepts} a sequence $s$ if and only if $\hat\delta(q_0, s) \in F$. Each DFA $\mathcal{D}$ hence represents a language $L(\mathcal{D}) = \{ s \mid \hat\delta(q_0,s) \in F \}$. Famously, the class of languages that a DFA can accept is precisely the set of regular languages. 

\emph{Moore machines} generalise DFAs by replacing the set of accepting states by an output function $\gamma: Q \to \Gamma$, with $\Gamma$ the output alphabet. Whereas DFAs are limited to $\Gamma = \{ \text{Reject}, \text{Accept} \}$, Moore machines allow an arbitrary number of labels, analogous to the distinction between binary and multiclass classification.

\subsubsection{Example Languages} \label{sec:example-languages}

\begin{figure}[tb]
    \centering
    \begin{subfigure}[t]{0.30\textwidth}
        \centering
        \includegraphics[width=0.9\textwidth]{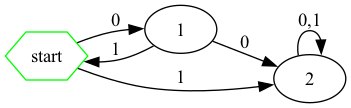}
        \caption{$\mathcal{D}_1$}
        \label{fig:example-languages-d1}
    \end{subfigure}
    \begin{subfigure}[t]{0.30\textwidth}
        \centering
        \includegraphics[width=0.8\textwidth]{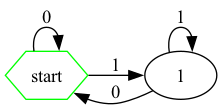}
        \label{fig:example-languages-g2}
        \caption{$\mathcal{G}_2$}
    \end{subfigure}
    \begin{subfigure}[t]{0.30\textwidth}
        \centering
        \includegraphics[width=0.8\textwidth]{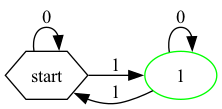}
        \caption{Parity}
        \label{fig:example-languages-parity}
    \end{subfigure}
    \caption{The DFAs accepting $\mathcal{D}_1$, $\mathcal{G}_2$ and Parity.}
    \label{fig:example-languages}
\end{figure}

We consider the following regular languages defined over the binary alphabet $\Sigma = \{0,1\}$. Figure \ref{fig:example-languages} shows some of their corresponding DFAs. All these languages are star-free except for Parity. 
\begin{itemize}
    \item \textbf{Ones}: The language Ones ($1^*$) contains all sequences consisting of only ones.
    \item \textbf{First}: The language First ($1(0\mid1)^*$) contains all sequences starting in one.
    \item\textbf{Depth-bounded Dyck ($\mathcal{D}_i$)}:
    The depth-bounded Dyck languages contain all sequences where the two symbols are correctly balanced and the maximal nesting depth is bounded. We denote the Dyck language with maximum depth $i$ as $\mathcal{D}_i$. For example, $\mathcal{D}_1 = (01)^*$ and $\mathcal{D}_2 = (0(01)^*1)^*$. 
    \item \textbf{Grid ($\mathcal{G}_n$)}: The DFAs accepting the Grid languages model a one-dimensional world with $n$ squares, see Figure \ref{fig:example-languages}. The two symbols are interpreted as moving one square/state to the left or right, such that trying to move beyond the boundaries results in staying on that square.
    \item \textbf{Parity}: Parity contains all sequences with an odd number of ones.
\end{itemize}

\section{Training Tasks} \label{sec:generalised-training-task}

Various training tasks have been used to study the trainability of regular languages. As such, the aim is often not to predict membership to a language directly. Rather it is to predict some task-specific labelling over the states of the automaton recognising that language. Such a labelling over states is equivalent to a Moore machine. We briefly introduce some common tasks.

In \emph{state prediction} \cite{liu2023transformers}, the transformer predicts the automaton's state after processing a sequence $s$. In other words, it predicts the generalised transition function $\hat\delta(q_0, s)$. Directly accepting a language, or \emph{membership prediction} \cite{yao2021self-membership}, comes down to predicting a binary label for each state indicating whether it is accepting.

\begin{figure}
    \centering
    \begin{subfigure}[t]{0.30\textwidth}
        \includegraphics[width=\linewidth]{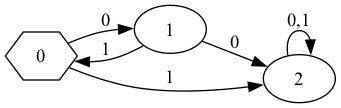}
    \caption{State prediction}
    \end{subfigure}
        \begin{subfigure}[t]{0.30\textwidth}
        \includegraphics[width=\linewidth]{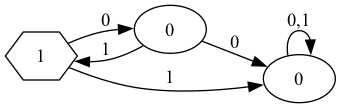}
    \caption{Membership prediction}
    \end{subfigure}
        \begin{subfigure}[t]{0.30\textwidth}
        \includegraphics[width=\linewidth]{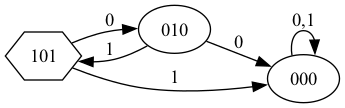}
    \caption{Character prediction}
    \end{subfigure}
    \caption{Labellings of $\mathcal{D}_1$ corresponding to the three tasks. The character prediction labels are encoded in binary. The three bits, from left to right, indicate the validity of $0$, $1$ and $\omega$.}
    \label{fig:labellings}
\end{figure}

A third example is \emph{character prediction} \cite{bhattamishra2020ability}, where the aim is to predict which next symbols are valid continuations of a sequence. A symbol $\sigma \in \Sigma$ is a valid continuation of a sequence $s$ if the extended sequence $s.\sigma$ has an accepting suffix, that is, $\exists s' \in \Sigma^*: s.\sigma.s' \in L$. Additionally, the special end-of-sequence symbol $\omega$ is a valid continuation of any accepting sequence. As the transformer must decide the validity of $\lvert \Sigma \cup \{\omega\}\rvert$ symbols, this is equivalent to predicting one of $2^{\lvert \Sigma \rvert+1}$ possible labels for each state. Figure \ref{fig:labellings} shows the labellings associated to these three tasks for the language $\mathcal{D}_1$.

Since state prediction corresponds to a unique labelling of each state, it corresponds to the fully observable case of predicting the automaton's state. In contrast, the other two tasks allow states to share a label, making them partially observable variants of this problem. In the end, they all boil down to the general task of simulating a Moore machine specific to the task and target language.

In addition to the training task, we highlight two other common choices in the training setup. Often \emph{sequence accuracy} is used in combination with character prediction \cite{bhattamishra2020ability,schmidhuber2001lstmRec,suzgun2019evaluating}. This metric counts correctly accepted sequences and considers a sequence accepted if all symbols in it are predicted to be valid and $\omega$ is valid in the final position. Furthermore, some studies train only on positive sequences \cite{bhattamishra2020ability,suzgun2019evaluating}, resembling the language modelling setting.

\section{Extracting State Machines}\label{sec:method-extracting}
Assuming a neural network faithfully implements an automaton, it must be possible to extract said automaton. Weiss et al. \cite{weiss2018extracting} proposed a method for extracting DFAs from recurrent neural networks (RNN) using the exact learning algorithm $L^*$ \cite{angluin1987lstar}. We briefly summarise their method below.

The $L^*$ algorithm learns a DFA by asking queries to a teacher, here the RNN. These queries are either membership queries, asking whether the teacher accepts a sequence, or equivalence queries, asking whether a proposed DFA implements the teacher's language and returning a counterexample should they differ. 

Learning a minimal DFA consistent with a set of positive and negative examples is NP-hard in general \cite{gold1978complexity}. However, $L^*$  can learn any minimal DFA in time polynomial in the number of states of that DFA from a teacher answering only membership and equivalence queries \cite{angluin1987lstar}.

Asking membership queries to a neural network is trivial, given that they accept the query as input and are trained as a classifier. Answering equivalence queries is generally intractable, however. Weiss et al. \cite{weiss2018extracting} deal with this intractability by checking equivalence against a discrete approximation of the RNN. They partition the hidden activation space such that each partition represents a state and state transitions are defined by a single sample in each partition \cite{giles1991extracting}. In other words, they keep track of two DFAs: the DFA $\mathcal{D}$ being extracted by $L^*$, and the DFA $\mathcal{D'}$ specified by the partitioning of the activation space. 

To answer the equivalence query, the states of $\mathcal{D}$ and $\mathcal{D}'$ are traversed breadth-first in parallel. Two types of conflicts may occur. First, the sequences accepted by $\mathcal{D}$ may differ from those accepted by the RNN, which produces a counterexample. Second, one state of $\mathcal{D}'$ may be associated with two different states of $\mathcal{D}$, in which case the partitioning must be refined. 

A decision tree keeps track of the partitioning, where internal nodes are support vector machines and leaves are states of $\mathcal{D}'$. A new support vector machine splits a leaf in this tree when a refinement is needed. 

A key strength of this partitioning approach is that it aligns the automaton’s states with discrete regions of the RNN's state space, ensuring the automaton reflects the RNN's continuous state representations.

\subsection{Extracting from Transformers} \label{sec:extracting-from-transformers}
Importantly, the method of Weiss et al. \cite{weiss2018extracting} is not specific to RNNs but can be used on any \textit{neural membership predictor}, provided a suitable state space is chosen. 
In a transformer, we propose using the activations just before unembedding. Similar to the state vectors of the RNNs in the original study, the output is computed from these activations with a single affine transformation followed by a softmax.  

This transformation from the state space to the output space represents the output function of the simulated DFA. For RNNs, the transition function of the DFA is correspondingly implemented in its recurrent connections and input function. However, for a transformer, this cannot hold as there are no recurrent connections. Instead, it is more accurate to think of the transformer's forward projection from a sequence to the chosen state space as simulating the \emph{generalised} transition function. This does make sense, as all previous symbols may impact the state directly through the attention mechanism.

\subsection{Extracting Moore Machines}
\label{sec:extracting-moore-machines}

On transformers not trained as membership predictors, we cannot directly apply the extraction method previously described. However, as covered in Section \ref{sec:generalised-training-task}, some of the alternative training tasks map to a labelling over the automaton's states. As a Moore machine is the formal description of such a labelling we cover all these tasks by extending the method to extract Moore machines.

To extract Moore machines instead of DFAs, we use a generalisation of $L^*$. This generalisation is a special case of an existing extension of $L^*$ that learns product Moore machines \cite{moerman2019product}. We briefly describe this special case here.

Internally, the learner keeps track of an observation table $\mathcal{O}$, where answers to membership queries are stored and from which proposals are constructed.

\begin{definition} (Observation Table)
  An \textit{observation table} over alphabet $\Sigma$ is a triple $\mathcal{O} = (S,E,T)$, with:
  \begin{itemize}
    \item $S$ a non-empty finite set of sequences, where $\forall w \in S: \text{prefix}(w) \in S$. \\$S$ is said to be \textit{prefix-closed}.
    \item $E$ a non-empty finite set of sequences, where $\forall w \in E: \text{suffix}(w) \in E$. \\$E$ is said to be \textit{suffix-closed}.
    \item $T$ a function mapping $((S \cup S . \Sigma) . E)$ to $\Gamma$.
  \end{itemize}
\end{definition}

The observation table can be thought of as a table where the rows are labelled with elements from $(S \cup S . \Sigma)$ and the columns with elements from $E$. The entry for row $s$ and column $e$ is then $T(s . e)$. The observation table defines a Moore machine $M(S,E,T)$ as $Q = \{row(s) \mid s \in S\}$, $\delta(row(s), \sigma) = row(s . \sigma)$, $q_0 = row(\epsilon)$, and $\gamma(s) = T(s)$.

An observation table is called \textit{closed} if, for each $t \in S.\Sigma$, there is an $s \in S$ such that $row(t) = row(s)$.
An observation table is called \textit{consistent} if for all $s_1, s_2 \in S$ and $\sigma \in \Sigma$, $row(s_1) = row(s_2) \Rightarrow row(s_1 . \sigma) = row(s_2 . \sigma)$. The following property follows from Moerman \cite{moerman2019product}.

\begin{theorem}
  If $(S,E,T)$ is a closed, consistent observation table, the Moore machine $M(S,E,T)$ is consistent with the finite function T.
\end{theorem}

\noindent
Finally, the complete algorithm consists of the following steps.
\begin{enumerate}
    \item Ask membership queries to make the (initially empty) observation table closed and consistent.
    \item Query the equivalence of $M(S,E,T)$ and the teacher's language.
    \item If equivalent, return $M(S,E,T)$; otherwise, extend T with the returned counterexample and return to step 1.
\end{enumerate}

\section{Related Work}
 Following the $L^*$ algorithm \cite{angluin1987lstar} for learning DFAs, more algorithms based on Angluin's exact learning framework \cite{Angluin1988framework} of using queries and counterexamples have been developed to learn other concepts, such as weighted finite automata \cite{balle2015extractingWFA} and horn theories \cite{Angluin1992horn}. Similar to Weiss et al. \cite{weiss2018extracting}, these have been applied to construct a high-level abstraction of neural networks \cite{weiss2019extractingWDFA,blum2024hornenvelopes}. Our work contributes to this line of research by extracting Moore machines from transformers. 
 
 Concurrently to our work, Zhang et al. \cite{zhang2024automataextractiontransformers} attempted to extract DFAs from transformers. As they claim transformers are inherently stateless, Zhang et al. \cite{zhang2024automataextractiontransformers} first train an RNN to simulate the transformer. Next, they apply the original extraction method by Weiss et al. \cite{weiss2018extracting} on that RNN. As now the discretisation is performed on the internal state of this intermediate RNN, the extracted DFA is not aligned with the original transformer but with that of the intermediate RNN. As discussed in Section \ref{sec:extracting-from-transformers}, transformers must simulate a state if they are to successfully recognise a regular language, which they can do by modelling the \emph{generalised} transition function, making the intermediate RNN redundant.
 
\section{Experiments}
In this section, we apply our extraction method to transformers trained on regular languages and analyse the resulting Moore machines to answer the following two research questions. All code to replicate the experiments is available on \url{https://github.com/Rik-A3/Extracting-Moore}.

\subsubsection{RQ1.} \textit{Do the extracted state machines faithfully model the language learnt by the transformer?}
There are two reasons why the extracted automaton could differ from the target language. Either because the transformer did not learn the correct language or because the extracted automaton does not correctly model the transformer. For the extraction method to be faithful, the latter reason should not occur.

\subsubsection{RQ2.} \textit{Can automata extraction help explain the impact of different training setups?} Two aspects of the training setup are studied more closely using the extraction technique. First, we investigate whether transformers can learn a language using only positive examples. Second, we look at sequence accuracy commonly used in combination with character prediction.

\subsection{Experiment Setup}

\subsubsection{Datasets.} The training datasets contain 10\,000 sequences of length 32 labelled using a target Moore machine. The validation and test sets contain 1\,000 sequences of length 100. Each sequence is prepended with a unique beginning-of-sequence symbol, allowing the transformer to learn an output for the start state. Unless stated otherwise, the sequences are generated by randomly traversing the target Moore machine. To prevent overly skewed class labels due to garbage states, the probability $p(q, \sigma)$ of picking the next symbol $\sigma$ in the state $q$ is taken as $p(q,0) = \frac{c(\delta(q,1))}{c(\delta(q,0)) + c(\delta(q,1) }$ for $\sigma=0$ and similarly for $\sigma=1$, where $c(q)$ is the number of times state $q$ has been visited during the generation of the dataset.

\subsubsection{Architecture.} We use one-layer encoder-only transformers with soft attention \cite{vaswani2017attention}, pre-norm layer normalisation \cite{nguyen2019prenorm}, and rotary positional encodings \cite{jianlin2023roformer}. The latter two techniques have been demonstrated to enhance the original transformer architecture in terms of training efficiency and length generalisation, respectively. We use a residual stream width of 16, a single-layer MLP block of width 64, and 4 attention heads. This is a subset of the transformer configurations studied in Bhattamishra et al. \cite{bhattamishra2020ability}. We do not compare different transformer configurations, as this has been extensively done before \cite{bhattamishra2020ability,deletang2023neural}.

\newpage
\subsubsection{Training.} We train with the Adam optimiser \cite{kingma2017adam} using a learning rate of $3 \cdot 10^{-4}$, no learning rate schedule, and a batch size of 32. We employ early stopping on the validation loss, with a patience of 5 epochs. Each experiment is repeated three times with different seeds. Training can be replicated in under six hours on a single NVIDIA GeForce RTX 3080 Ti GPU.

\subsubsection{Extraction.} Weiss et al. \cite{weiss2018extracting} employ three tactics to improve practical performance, which we also adopt. First, we impose a time limit after which the algorithm is stopped and the latest automaton proposed during an equivalence query is returned. Second, we use an aggressive initial splitting strategy on the first set of conflicting activation vectors by splitting along the dimensions where these vectors differ the most. The number of dimensions $d$ to initially split on is a hyperparameter called the \emph{initial split depth}, which we keep at 10. Third, all possible sequences of length three are supplied as starting examples for the $L^*$ learner. These last two tactics prevent the algorithm from terminating prematurely on an automaton with a single state.
 
\subsection{Are the extracted state machines faithful?}

To evaluate our approach, we trained transformers to recognise each of the example languages described in Section \ref{sec:example-languages} under the state prediction task, as this is the fully observable case. We measure the prediction accuracy on the final ten positions of sequences in the test set, emphasising generalisation beyond the training length. We also report the \emph{average correct length} (ACL), meaning the average length up to which the transformer makes no mistakes.

From each of these transformers, we extract a Moore machine. Again we report the prediction accuracy on the last ten positions and average correct length, where now the predictions of the transformer serve as the ground truth, which the extracted Moore machine must predict. All results are in Table \ref{tab:main}. When the extracted automaton is exactly the target, all mistakes are due to the transformer imperfectly modelling the language. This is why we include how often the exact target machine was extracted in the last column.

\begin{table}[tb]
\caption{We report the accuracy and average correct length of transformers on a test set of length 100 as well as the accuracy and average correct length of the extracted Moore machine compared to predictions of the transformer on the same test set. Results are averaged over 3 runs.
 We report the extraction time, where T/O indicates a time-out of 30s, and how often the target Moore machine was extracted.}
\label{tab:main}
\centering
\begin{tabular}{lcccccc}
\toprule
& \multicolumn{2}{c}{Test} & \multicolumn{4}{c}{Extraction}  \\
\cmidrule(lr){2-3} \cmidrule(lr){4-7}
Name & Accuracy & ACL  & Accuracy & ACL & Time & Target? \\
\midrule
$First$ & 0.650 & 59.292 & 0.650 & 59.667 & 0.15, T/O, 0.19 & 2/3 \\
$Ones$ & 1.000 & 101.000 & 1.000 & 101.000 & 0.10, 0.11, 0.13 & 3/3 \\
$\mathcal{G}_2$ & 1.000 & 101.000 & 1.000 & 101.000 & 0.16, 0.15, 0.16 & 3/3 \\
$\mathcal{G}_3$ & 0.950 & 77.125 & 0.950 & 77.125 & 0.21, 0.17, 0.20 & 3/3 \\
$\mathcal{D}_1$ & 0.996 & 88.125 & 0.996 & 88.125 & 0.20, 0.25, 0.19 & 3/3 \\
$\mathcal{D}_2$ & 1.000 & 44.958 & 1.000 & 44.958 & 0.22, 0.23, 0.30 & 3/3 \\
$Parity$ & 0.458 & 17.083 & 0.496 & 17.917 & 0.24, 0.12, T/O & 2/3 \\
\bottomrule
\end{tabular}
\end{table}

The languages First, Ones, $\mathcal{G}_2$, $\mathcal{G}_3$, $\mathcal{D}_1$ and $\mathcal{D}_2$ are all learnable by our transformers as they achieve high accuracy and generalise well beyond the training length of 32. A single transformer failed to learn the First language well, potentially due to suboptimal hyperparameters. For Parity, generalisation does not occur. Only sequences shorter than the training length are predicted accurately.  

The extraction results show that the target automaton is often recovered from the transformer. This is what we expect when the transformer has perfectly learnt the language beyond the length checked during the breadth-first search. We thus conclude our method finds a small and faithful automaton resembling the language accepted by the transformer. However, when the transformer's modelling does not generalise well towards longer sequences, the abstraction inevitably breaks down, and our method returns a large automaton when the time limit is reached. This intolerance to noise is typical for exact learning algorithms such as $L^*$. Although large, the returned intermediate Moore machine does represent the transformer slightly better than the target machine, as shown by the increased accuracy and average correct length on the test set.
\\

\noindent
\textbf{RQ1.} Our results show that the correct automata is extracted in all cases where the transformer successfully learnt the language.

\subsection{Can automata extraction explain experimental differences?} 

Similar to the previous section, we train transformers on the example languages, apply the extraction method, and report the average correct length as well as the accuracy on the last ten positions of the test set. However, we now consider two common variations on the training setup.

\subsubsection{Positive-only learning}
First, we train the transformers on a dataset containing only positive examples, similar to Bhattamishra et al. \cite{bhattamishra2020ability}. The dataset is generated by traversing the target DFA at random after removing the garbage state for efficiency and only keeping accepted sequences. The results are shown in Table \ref{tab:positive-only}. For languages $\mathcal{G}_2$, $\mathcal{G}_3$ and Parity, the results are as before. The other languages are learnt poorly. As we will explain, this is because of their garbage state.

\begin{table}[tb]
\caption{We report the accuracy and average correct length of positive-only trained transformers on a test set of length 100 as well as the accuracy and average correct length of the extracted Moore machine compared to predictions of the transformer on the same test set. Results are averaged over 3 runs.
 We furthermore report the extraction time, where T/O indicates a time-out of 30s, and how often the target Moore machine was extracted.}
\label{tab:positive-only}
\centering
\begin{tabular}{lcccccc}
\toprule
& \multicolumn{2}{c}{Test} & \multicolumn{4}{c}{Extraction}  \\
\cmidrule(lr){2-3} \cmidrule(lr){4-7}
Name & Accuracy & ACL  & Accuracy & ACL & Time & Target? \\
\midrule
$First$ & 0.417 & 42.667 & 1.000 & 101.000 & 0.18, 0.15, 0.15 & 0/3\\
$Ones$ & 0.000 & 10.167 & 1.000 & 101.000 & 0.06, 0.05, 0.06 & 0/3 \\
$\mathcal{G}_2$ & 1.000 & 101.000 & 1.000 & 101.000 & 0.15, 0.27, 0.13 & 3/3 \\
$\mathcal{G}_3$ & 0.992 & 93.750 & 0.992 & 93.750 & 0.19, 0.16, 0.16 & 3/3 \\
$\mathcal{D}_1$ & 0.000 & 17.000 & 1.000 & 97.000 & 0.10, 0.11, T/O & 2/3 \\
$\mathcal{D}_2$ & 0.000 & 10.583 & 0.621 & 35.333 & T/O, T/O, T/O & 0/3 \\
$Parity$ & 0.471 & 15.458 & 0.517 & 15.917 & T/O, 0.12, 0.11 & 2/3 \\
\bottomrule
\end{tabular}
\end{table}

Since the gradient update of a transformer is not only based on the prediction of the last symbol of the sequence but of all symbols in the sequence, the transformer is essentially trained on all prefixes of the examples in the training dataset. When there is no garbage state, a positive-only dataset containing accepting examples of a certain length can have in its prefix set all sequences---including non-accepting ones---of a length slightly smaller than the dataset length. When there is a garbage state, however, sequences visiting it never appear as the prefix of a positive example and are, therefore, never seen by the transformer.

\begin{figure}[t]
    \centering
    \begin{subfigure}[t]{0.30\textwidth}
        \centering
        \includegraphics[width=0.30\textwidth]{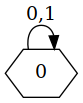}
        \caption{Ones}
    \end{subfigure}
    \begin{subfigure}[t]{0.30\textwidth}
        \centering
        \includegraphics[width=0.85\textwidth]{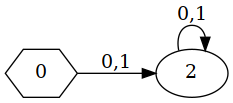}
        \caption{First}
    \end{subfigure}
    \begin{subfigure}[t]{0.30\textwidth}
        \centering
        \includegraphics[width=0.75\textwidth]{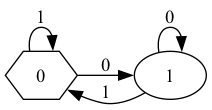}
        \caption{$\mathcal{D}_1$}
    \end{subfigure}
    \caption{Moore machines extracted from transformers trained on Ones, First, and $\mathcal{D}_1$.}
    \label{fig:pos-only-extraction}
\end{figure}

We thus expect that garbage states will not be learnt and only languages containing a garbage state will be affected. This explains the decreased scores on a test set containing both positive and negative examples. Nonetheless, the extraction method finds small, faithful Moore machines for Ones, First, and $\mathcal{D}_1$. These automata, shown in Figure \ref{fig:pos-only-extraction}, are exactly the target automata where the garbage state is not modelled. 

This observation is related to recent work \cite{hahn2024sensitive} showing encoder-only transformers are biased towards learning \textit{functions of low sensitivity}, that is, functions where flipping a single bit in the input often does not cause a change in the output. All three alternative languages the transformer has learnt are less sensitive than the expected languages with a garbage state, while still accepting all examples in the dataset. We leave further investigations on the sensitivity of languages with or without garbage states to future work.

\newpage
\subsubsection{Character prediction and sequence accuracy}
This observation sparks a discussion on the use of another common training setup: character prediction and sequence accuracy. In the character prediction task, a symbol in a sequence is invalid if from that point no accepting suffixes exist, meaning the sequence is in a garbage state. Since, according to sequence accuracy, a sequence is rejected when a single symbol is predicted to be invalid, all predictions of the symbols after it are irrelevant. Therefore,  transformers can achieve perfect sequence accuracy by correctly labelling non-garbage states without explicitly modelling the garbage state. Indeed, the automaton for language $\mathcal{G}_{n}$, given the correct labelling, would score a perfect sequence accuracy on sequences labelled using $\mathcal{D}_{n-1}$.  \\

\noindent
\textbf{RQ2}: Using our approach, we have shown transformers do not learn garbage states when trained on only positive examples and observe that the sequence accuracy can be high even though a language different from the target has been learnt. These examples highlight that judging whether a language has been learnt based on simple metrics can lead to misleading results, which our method prevents by extracting said language.
 
\section{Conclusion}
Inspired by prior work leveraging Angluin's exact learning framework to learn high-level abstractions from neural networks, we extracted Moore machines from encoder-only transformers trained on regular languages using an extension of $L^*$. Our method accommodates a variety of training tasks used in literature since, as we have shown, many of these tasks reduce to simulating Moore machines. We demonstrated that our extraction technique finds small automata that faithfully describe the transformer, given that the transformer learnt the language. Finally, we found that transformers do not model garbage states when trained on only positive examples and that the common sequence accuracy can report a perfect score even though the transformer has learnt a language different from the target.

\begin{credits}
\subsubsection{\ackname}
This research received funding from the Flemish Government (AI Research Program), the Flanders Research Foundation (FWO) under project G097720N, and the European Research Council (ERC) under the European Union’s Horizon 2020 research and innovation program (Advanced Grant DeepLog No. 101142702). We are grateful to Luc De Raedt for his helpful feedback.\\
\end{credits}

\begin{credits}
\subsubsection{\discintname} The authors have no competing interests to declare that are relevant to the content of this article.
\end{credits}

\bibliography{references}

\begin{thebibliography}{10}
\providecommand{\url}[1]{\texttt{#1}}
\providecommand{\urlprefix}{URL }
\providecommand{\doi}[1]{https://doi.org/#1}

\bibitem{ackerman2020survey}
Ackerman, J., Cybenko, G.: A survey of neural networks and formal languages. arXiv preprint arXiv:2006.01338  (2020)

\bibitem{angluin1987lstar}
Angluin, D.: Learning regular sets from queries and counterexamples. Information and computation  \textbf{75}(2),  87--106 (1987)

\bibitem{Angluin1988framework}
Angluin, D.: Queries and concept learning. Machine Learning  \textbf{2}(4),  319--342 (Apr 1988). \doi{10.1023/A:1022821128753}

\bibitem{Angluin1992horn}
Angluin, D., Frazier, M., Pitt, L.: Learning conjunctions of horn clauses. Machine Learning  \textbf{9}(2),  147--164 (Jul 1992)

\bibitem{anil2022exploring}
Anil, C., Wu, Y., Andreassen, A., Lewkowycz, A., Misra, V., Ramasesh, V., Slone, A., Gur-Ari, G., Dyer, E., Neyshabur, B.: Exploring length generalization in large language models. Advances in Neural Information Processing Systems  \textbf{35},  38546--38556 (2022)

\bibitem{balle2015extractingWFA}
Balle, B., Mohri, M.: Learning weighted automata. pp. 1--21 (09 2015)

\bibitem{barrington1992regular}
Barrington, D.A.M., Compton, K., Straubing, H., Th{\'e}rien, D.: Regular languages in nc1. Journal of Computer and System Sciences  \textbf{44}(3),  478--499 (1992)

\bibitem{bhattamishra2020ability}
Bhattamishra, S., Ahuja, K., Goyal, N.: On the {A}bility and {L}imitations of {T}ransformers to {R}ecognize {F}ormal {L}anguages. In: Proceedings of the 2020 Conference on Empirical Methods in Natural Language Processing (EMNLP). pp. 7096--7116. Association for Computational Linguistics (2020)

\bibitem{blum2024hornenvelopes}
Blum, S., Koudijs, R., Ozaki, A., Touileb, S.: Learning horn envelopes via queries from language models. International Journal of Approximate Reasoning  \textbf{171},  109026 (2024), synergies between Machine Learning and Reasoning

\bibitem{chiang2022overcoming}
Chiang, D., Cholak, P.: Overcoming a theoretical limitation of self-attention. In: Proceedings of the 60th Annual Meeting of the Association for Computational Linguistics (2022)

\bibitem{deletang2023neural}
Delétang, G., Ruoss, A., Grau-Moya, J., Genewein, T., Wenliang, L.K., Catt, E., Cundy, C., Hutter, M., Legg, S., Veness, J., Ortega, P.A.: Neural networks and the chomsky hierarchy. The Eleventh International Conference on Learning Representations  (2022)

\bibitem{schmidhuber2001lstmRec}
Gers, F., Schmidhuber, E.: Lstm recurrent networks learn simple context-free and context-sensitive languages. IEEE Transactions on Neural Networks  \textbf{12}(6),  1333--1340 (2001)

\bibitem{giles1991extracting}
Giles, C.L., Miller, C.B., Chen, D., Sun, G.Z., Chen, H.H., Lee, Y.C.: Extracting and learning an unknown grammar with recurrent neural networks. In: Moody, J., Hanson, S., Lippmann, R. (eds.) Advances in Neural Information Processing Systems. vol.~4. Morgan-Kaufmann (1991)

\bibitem{gold1978complexity}
Gold, E.M.: Complexity of automaton identification from given data. Information and Control  \textbf{37}(3),  302--320 (1978). \doi{https://doi.org/10.1016/S0019-9958(78)90562-4}

\bibitem{hahn2024sensitive}
Hahn, M., Rofin, M.: Why are sensitive functions hard for transformers? In: Ku, L.W., Martins, A., Srikumar, V. (eds.) Proceedings of the 62nd Annual Meeting of the Association for Computational Linguistics (Volume 1: Long Papers). pp. 14973--15008. Association for Computational Linguistics, Bangkok, Thailand (Aug 2024). \doi{10.18653/v1/2024.acl-long.800}

\bibitem{hopcroft2006automata}
Hopcroft, J.E., Motwani, R., Ullman, J.D.: Introduction to Automata Theory, Languages, and Computation (3rd Edition). Addison-Wesley Longman Publishing Co., Inc., USA (2006)

\bibitem{kingma2017adam}
Kingma, D.P., Ba, J.L.: Adam: A method for stochastic gradient descent. In: International Conference on Learning Representations. pp. 1--15 (2015)

\bibitem{liu2023transformers}
Liu, B., Ash, J.T., Goel, S., Krishnamurthy, A., Zhang, C.: Transformers learn shortcuts to automata. In: The Eleventh International Conference on Learning Representations (2023)

\bibitem{merrill2021formal}
Merrill, W.: Formal language theory meets modern nlp. arXiv preprint arXiv:2102.10094  (2021)

\bibitem{moerman2019product}
Moerman, J.: Learning product automata. In: Unold, O., Dyrka, W., Wieczorek, W. (eds.) Proceedings of Machine Learning Research. vol.~93, pp. 54--66. PMLR (2019)

\bibitem{nguyen2019prenorm}
Nguyen, T.Q., Salazar, J.: Transformers without tears: Improving the normalization of self-attention. In: Niehues, J., Cattoni, R., St{\"u}ker, S., Negri, M., Turchi, M., Ha, T.L., Salesky, E., Sanabria, R., Barrault, L., Specia, L., Federico, M. (eds.) Proceedings of the 16th International Conference on Spoken Language Translation. Association for Computational Linguistics, Hong Kong (Nov 2-3 2019)

\bibitem{Strobl2023TransformersAR}
Strobl, L., Merrill, W., Weiss, G., Chiang, D., Angluin, D.: Transformers as recognizers of formal languages: A survey on expressivity (2023)

\bibitem{jianlin2023roformer}
Su, J., Ahmed, M., Lu, Y., Pan, S., Bo, W., Liu, Y.: Roformer: Enhanced transformer with rotary position embedding. Neurocomputing  \textbf{568},  127063 (2024)

\bibitem{suzgun2019evaluating}
Suzgun, M., Belinkov, Y., Shieber, S.M.: On evaluating the generalization of lstm models in formal languages. pp. 277--286 (January 2019)

\bibitem{vaswani2017attention}
Vaswani, A., Shazeer, N., Parmar, N., Uszkoreit, J., Jones, L., Gomez, A.N., Kaiser, L.u., Polosukhin, I.: Attention is all you need. In: Guyon, I., Luxburg, U.V., Bengio, S., Wallach, H., Fergus, R., Vishwanathan, S., Garnett, R. (eds.) Advances in Neural Information Processing Systems. vol.~30. Curran Associates, Inc. (2017)

\bibitem{weiss2018extracting}
Weiss, G., Goldberg, Y., Yahav, E.: Extracting automata from recurrent neural networks using queries and counterexamples. In: International Conference on Machine Learning. pp. 5247--5256. PMLR (2018)

\bibitem{weiss2019extractingWDFA}
Weiss, G., Goldberg, Y., Yahav, E.: Learning deterministic weighted automata with queries and counterexamples. In: Wallach, H., Larochelle, H., Beygelzimer, A., d\textquotesingle Alch\'{e}-Buc, F., Fox, E., Garnett, R. (eds.) Advances in Neural Information Processing Systems. vol.~32. Curran Associates, Inc. (2019)

\bibitem{yang2023maskedht}
Yang, A., Chiang, D., Angluin, D.: Masked hard-attention transformers recognize exactly the star-free languages (2023)

\bibitem{yao2021self-membership}
Yao, S., Peng, B., Papadimitriou, C., Narasimhan, K.: Self-attention networks can process bounded hierarchical languages. In: Zong, C., Xia, F., Li, W., Navigli, R. (eds.) Proceedings of the 59th Annual Meeting of the Association for Computational Linguistics and the 11th International Joint Conference on Natural Language Processing (Volume 1: Long Papers). pp. 3770--3785. Association for Computational Linguistics, Online (Aug 2021)

\bibitem{zhang2024automataextractiontransformers}
Zhang, Y., Wei, Z., Sun, M.: Automata extraction from transformers (2024)

\end{thebibliography}
\bibliographystyle{splncs04}

\end{document}